\documentclass[final]{egbib}

\usepackage{times}
\usepackage{epsfig}
\usepackage{graphicx}
\usepackage{amsmath}
\usepackage{amssymb}

\usepackage{subfig}
\usepackage{comment}
\usepackage{multirow}
\usepackage{soul}
\usepackage{verbatim}
\usepackage{algorithm}
\usepackage{algorithmic}
\usepackage{amsmath}
\usepackage{amssymb}
\usepackage{dsfont}
\usepackage{bbm}
\usepackage{mathtools}

\usepackage[pagebackref=true,breaklinks=true,colorlinks,bookmarks=false]{hyperref}

\begin{document}

\title{Suppressing Spoof-irrelevant Factors for Domain-agnostic Face Anti-spoofing}

\author{Taewook Kim\\
Kakao Enterprise\\
{\tt\small taewook101@postech.ac.kr}
\and
Yonghyun Kim \\
Kakao Enterprise\\
{\tt\small aiden.d@kakaoenterprise.com}}

\maketitle

\begin{abstract}
Face anti-spoofing aims to prevent false authentications of face recognition systems by distinguishing whether an image is originated from a human face or a spoof medium. We propose a novel method called Doubly Adversarial Suppression Network (DASN) for domain-agnostic face anti-spoofing; DASN improves the generalization ability to unseen domains by learning to effectively suppress spoof-irrelevant factors (SiFs) (e.g., camera sensors, illuminations). To achieve our goal, we introduce two types of adversarial learning schemes. In the first adversarial learning scheme, multiple SiFs are suppressed by deploying multiple discrimination heads that are trained against an encoder. In the second adversarial learning scheme, each of the discrimination heads is also adversarially trained to suppress a spoof factor, and the group of the secondary spoof classifier and the encoder aims to intensify the spoof factor by overcoming the suppression. We evaluate the proposed method on four public benchmark datasets, and achieve remarkable evaluation results. The results demonstrate the effectiveness of the proposed method.

\end{abstract}

\section{Introduction}
\noindent
Computerized face-recognition techniques \cite{deng2019arcface,Kim_2020_CVPR,cosface,regularface} have been successfully deployed in a wide range of real-world applications, such as criminal identification and e-commerce systems.
Despite its remarkable accuracy, these face recognition systems can erroneously authenticate deceivers as genuine users if the facial images of the genuine users are displayed.
Therefore, to ensure the security of face recognition, it is important to devise a way to prevent this error.

\begin{figure}[t!]
	\begin{center}
		\includegraphics[width=1\linewidth]{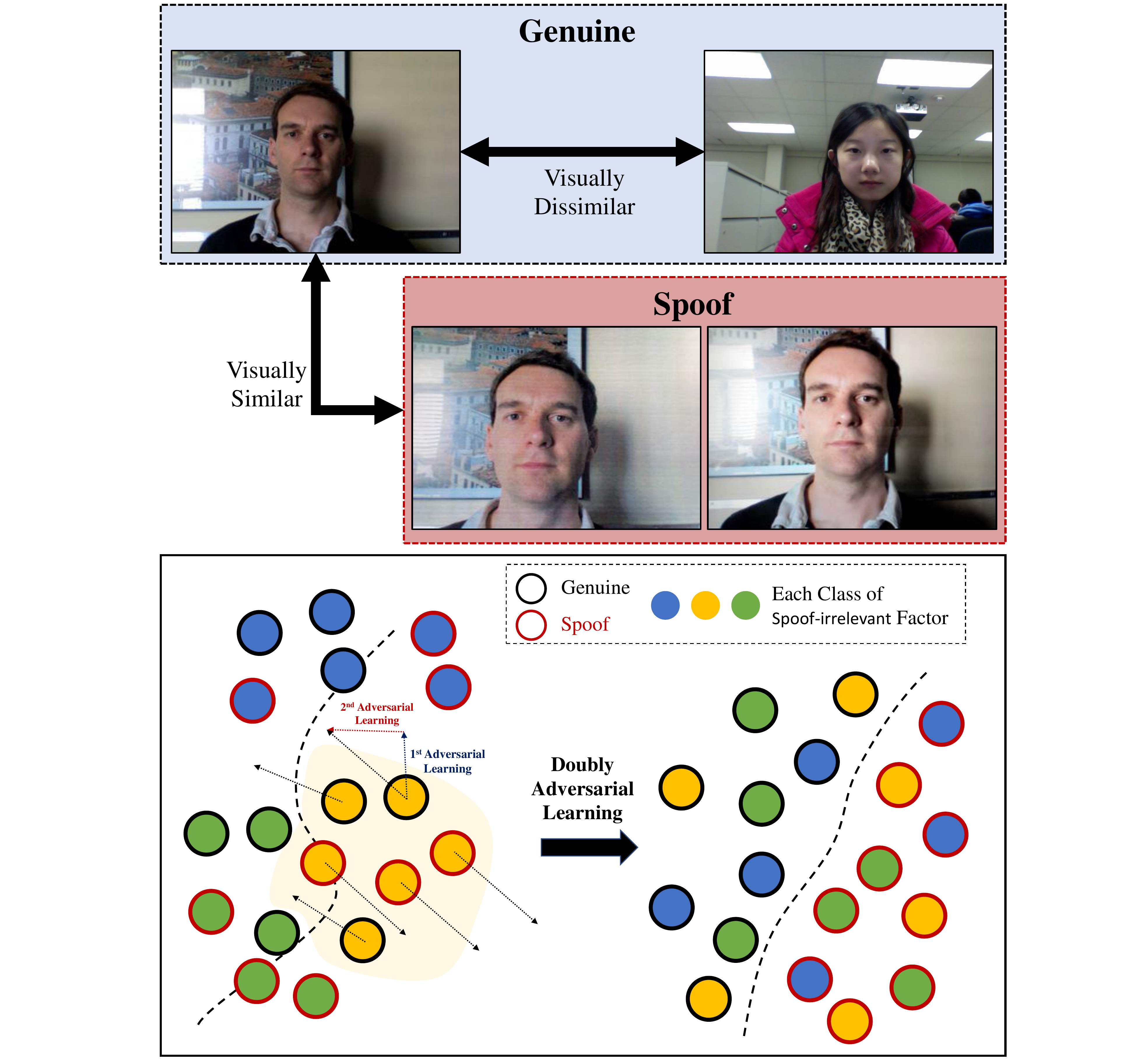}
	\end{center}
	\caption{\textbf{Top}: The spoof and genuine images that share the same SiFs, such as identity, background, and illuminations are visually similar to each other than the two genuine images that do not share the factors. 
	\textbf{Bottom}: 
	In our method, two types of adversarial learning strategies are deployed, hence our method is termed Doubly Adversarial Learning. The first adversarial learning is deployed to suppress SiFs, and samples that share the same SiFs are dispersed as a result. The second adversarial learning is deployed to intensify the spoof factor in the encoded features.
	}
	\label{fig:concept}
\end{figure}

Several methods have been proposed to prevent deceptions by capturing faithful spoof factors with the guidance of auxiliary supervisions. These methods utilized a priori knowledge to properly guide the network to capture generalizable cues. \cite{atoum2017face,liu2018learning} utilized the fact that spoof medium and genuine faces are distinctively different in terms of facial depth, and exploited pseudo-depth maps as auxiliary supervision. Other methods have been proposed to utilize reflection maps \cite{kim2019basn} or physiological signals \cite{liu2018learning} along with the facial depth maps, and achieved notable improvements.

Images captured in unconstrained real-world scenarios are largely diverse in terms of spoof-irrelevant factors, such as identities or background patterns, compared to the collected images of the existing datasets. Since it is impossible to collect images by considering every possible combination of these factors, large discrepancies between a training set and real-world data are inevitable. Therefore, it is desirable to mitigate the effect of such spoof-irrelevant factors, so that the model learns faithful spoof factors independently from spoof-irrelevant factors by using the training datasets that are limitedly diversified.

Here, the term \emph{spoof-irrelevant factors (SiFs)} indicates the factors that are uninformative and irrelevant to the face anti-spoofing, and at the same time induces visual similarities.
For instance, the spoof and genuine images of the same identity may look much similar to each other than the two images of the same class (e.g., spoof or genuine) with different identities.
Therefore, if features are na\"ively extracted from images without taking the SiFs into account, then the extracted features may be located in accordance with the SiFs, and deterioration of face anti-spoofing accuracy would be yielded (Figure \ref{fig:concept}).

Some methods \cite{Stehouwer_2020_CVPR,wang2020cross} have been proposed to increase generalization ability by alleviating the effects of SiFs. \cite{Stehouwer_2020_CVPR} proposed a method to train a model to achieve invariance to noise patterns incurred by sensory devices. \cite{wang2020cross} attempted to make a model to be invariant to identities by explicitly disentangling identity features using identity labels.
Multiple types of SiFs exist in the face images, however, the existing methods have considered only a single factor, hence not readily generalized. 

Our insight is that we need a learning scheme that efficiently suppresses multiple SiFs simultaneously in a way that does not disturb the learning of a spoof factor.
Various types of SiFs exist in images, thus reducing the effect of only a single SiF may erroneously guide the network to encode the remaining SiFs as spoof factors. In this regard, it is important that we design the model to be invariant to multiple SiFs.
In addition, we aim to build the model to stably suppress multiple SiFs, so that suppression of multiple SiFs does not dominate the overall learning procedure of the encoder, nor weakens the discriminative power of the encoder.
To realize our goal, we propose an architecture called Doubly Adversarial Suppression Network (DASN) that suppresses SiFs by adopting a doubly adversarial learning strategy. To consider various types of SiFs, we deploy multiple discrimination heads after an encoder, and adversarial learning is conducted between the discrimination heads and the encoder. 
Furthermore, our DASN performs additional adversarial learning between the group of the encoder and the secondary spoof classifier, and the intermediate layer of each discrimination head to intensify the spoof factor in the encoded feature. Incorporating the suppression of multiple SiFs is more challenging compared to the suppression of a single SiFs since the training process is hindered by instability. We note that by deploying the second adversarial learning scheme, the overall training process becomes stabilized and the performance is further boosted.
The two different adversarial learning procedures are performed comprehensively, hence the overall learning scheme is termed Doubly Adversarial Learning.

The contributions of our work can be summarized as follows:
\begin{itemize}
    \item We propose a DASN, which adopts doubly adversarial learning to effectively suppress the spoof-irrelevant factors, and intensify a spoof factor for enhanced generalization ability.
    \item DASN achieves the-state-of-the-art performance on various benchmark datasets \cite{oulu_npu,replay_attack,msu_fasd,casia_fasd} for domain generalization of face anti-spoofing. Moreover, our extensive ablation studies show that the suppression of SiFs is effectively conducted by adopting our DASN.
\end{itemize}


\section{Related Works}
\noindent{\textbf{Face Anti-Spoofing Methods.}} 
Face anti-spoofing is becoming increasingly important as concerns about the security of face-recognition systems grow. 
Many face anti-spoofing methods have been proposed; early methods used hand-crafted features, such as Local Binary Patterns (LBP) \cite{boulkenafet2015face,maatta2011face} or Histogram oriented Gradients (HoG) \cite{komulainen2013context,yang2013face} to solve the problem. 
After the success of deep learning in various tasks, many researchers proposed methods that use deep features \cite{li2016original,patel2016cross,yang2014learn}, and achieved improvements over the traditional methods. 
More recently, researchers considered auxiliary information, such as facial depth maps  \cite{atoum2017face}, physiological signals \cite{liu2018learning}, and reflection maps \cite{kim2019basn} to capture discriminative cues that could be universally applied to different domains. 
The researchers exploited the difference in these auxiliaries between spoof media and real humans (e.g., live humans exhibit physiological signals, whereas spoof mediums do not), and achieved notable achievements by showing improved generalization to unseen domains. 
However, the existing methods depend on the inaccurate pseudo-labels (e.g., depth maps are limited to facial areas), and the performance is still not satisfactory due to the large variations among different domains.
\smallbreak

\noindent{\textbf{Domain Generalization Methods.}} 
To address the problems, researchers aimed to tackle the face anti-spoofing in the perspective of domain-generalization to take advantage of multiple seen data \cite{saha2020domain,Shao_2019_CVPR,shao2020regularized}.
 \cite{Shao_2019_CVPR} proposed a multi-adversarial network that aims to enhance the generalization ability to unseen domains by training a generator to learn a feature space that is shared by multiple discriminators pre-trained on different source domains.
 \cite{shao2020regularized} proposed a method to supervise the network with generalized learning directions by incorporating domain shift scenarios in the meta-learning framework.
 \cite{saha2020domain} attempted to learn spatio-temporal features by deploying both the image-based network and video-based network, and utilized class-conditional domain discriminators to further improve the generalization capability. 
 \cite{wang2020cross} proposed a method that disentangles identity factors to be invariant to an identity.
However, \cite{wang2020cross} has considered only a single number of the spoof-irrelevant factors that can disturb the face anti-spoofing. In contrast, we fully consider all types of spoof-irrelevant factors that can be utilized.

\section{Proposed Method}
\begin{figure}[t!]
	\begin{center}
		\includegraphics[width=1\linewidth]{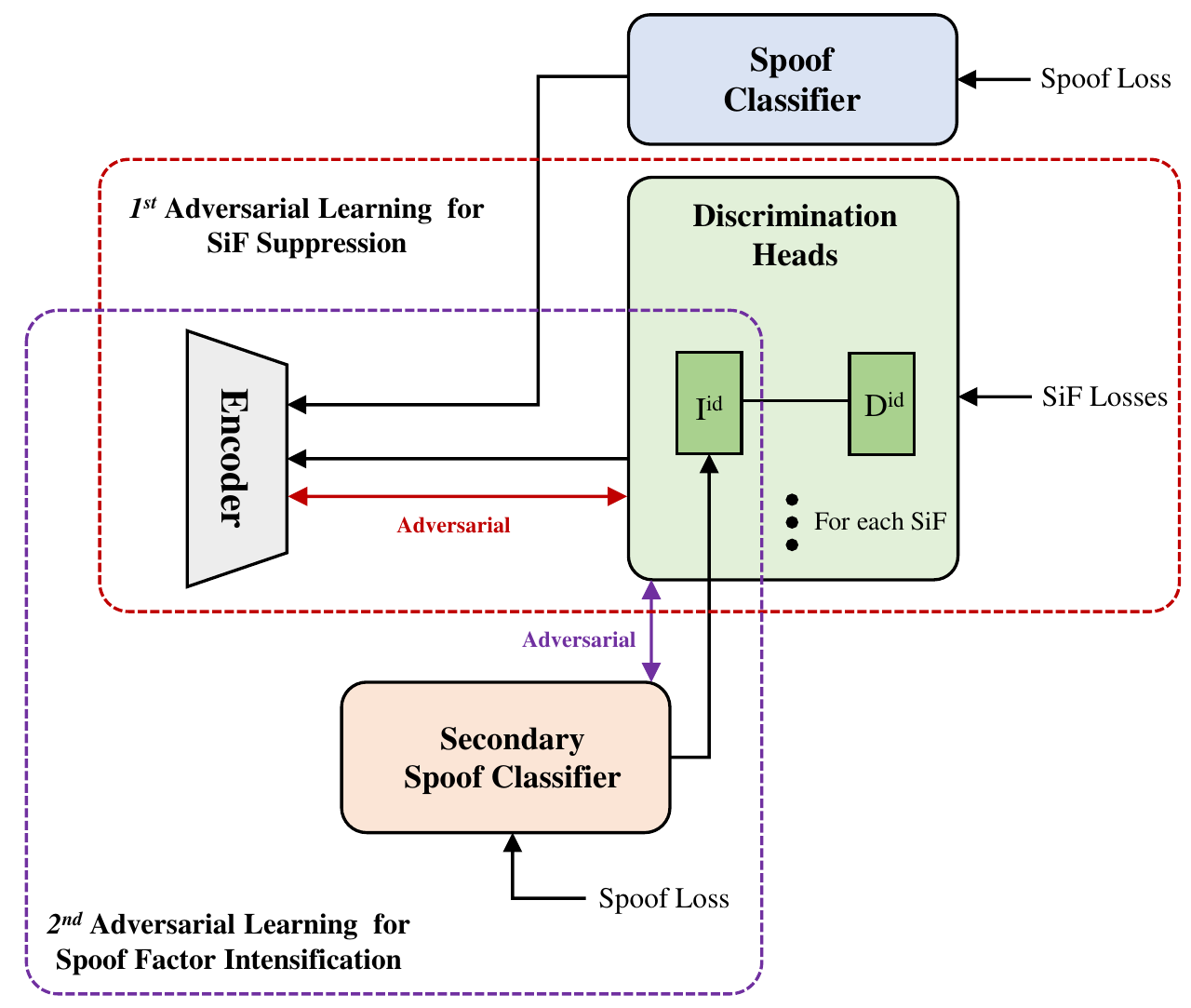}
	\end{center}
	\caption{
	Our doubly adversarial learning scheme; 
	(1) $1^{st}$ adversarial learning (red) between an encoder and the discrimination heads to suppress multiple SiFs,
	and 
	(2) $2^{nd}$ adversarial learning (purple) between the group of the encoder and the secondary spoof classifier, and SiF-aware intermediate layers to intensify the spoof factor.}
	\label{fig:method_overview}
\end{figure}
In this section, we first discuss the spoof-irrelevant factors. Then we introduce a doubly adversarial learning scheme for suppressing spoof-irrelevant factors and intensifying the spoof factor in the encoded features. Finally, we describe a Doubly Adversarial Suppression Network (DASN), which is a spoof classification network that is trained using the proposed doubly adversarial learning.

\begin{figure*}[t!]
	\begin{center}
		\includegraphics[width=1\linewidth]{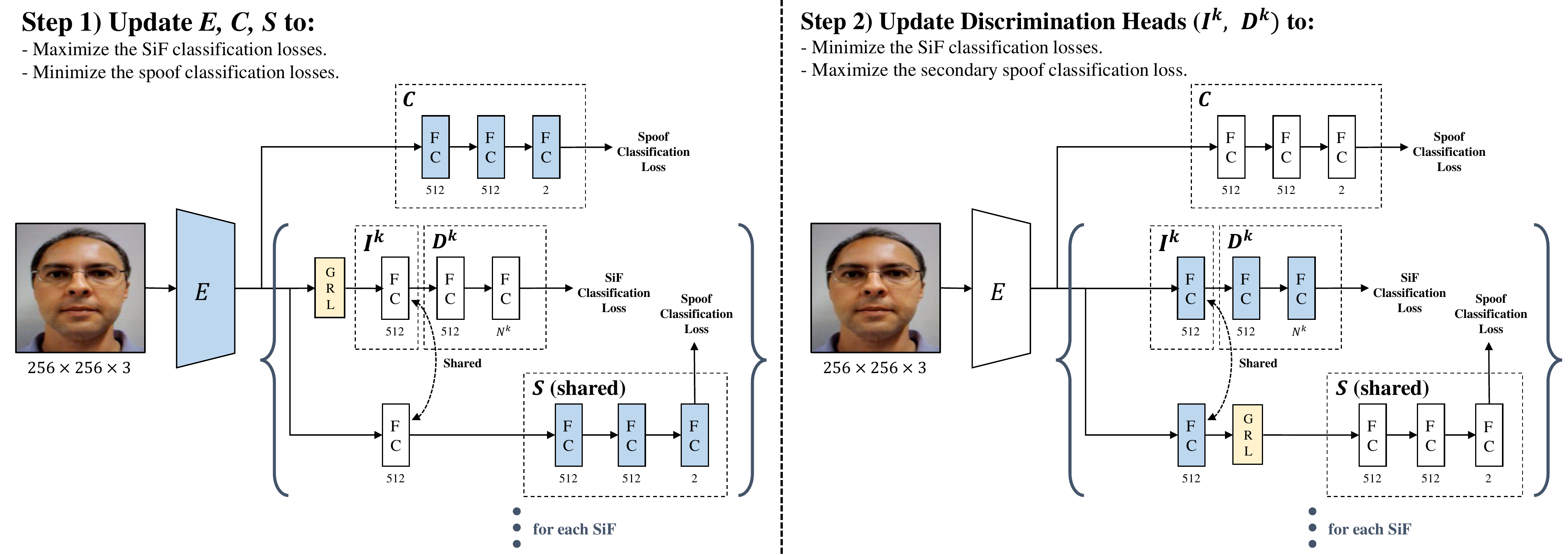}
	\end{center}
	\caption{
	Overview of DASN; the network is trained by our doubly adversarial learning in two step.
	In the first learning step, the encoder, the spoof classifier, and the secondary classifier strive to suppress SiFs and intensify the spoof factor in a collaborative way \cite{song2018collaborative}. A GRL reverses the sign of the gradients so that the encoder is updated in a way that suppresses SiFs. 
	In the second learning step, the discrimination heads learn to suppress the spoof factor and classify SiFs; the SiF-aware intermediate layer $I^{k}$ and the discriminator $D^{k}$ are updated with another GRL to suppress the spoof factor in $I^{k}$.
	In each stage, only the blue-colored parts are updated.
	As the training progresses, the discrimination heads gradually diverge, and DASN successfully suppresses the three SiFs: $k \in \{identity, environment, sensors\}$, and the spoof factor is intensified as the encoder overcomes the spoof factor suppression.
	}
	\label{fig:architecture}
\end{figure*}

\subsection{Spoof-irrelevant Factors}
We term the spoof-irrelevant factors as the factors that are irrelevant and uninformative to the face anti-spoofing, and that can incur visual similarity when the same type of SiFs are shared by images. 
For example, information regarding the facial structure of a subject, or gender is not meaningful in terms of detecting face spoofs, and different images of the same identity would be visually alike to each other (Figure \ref{fig:concept}). Hence, the identity of a face is an irrelevant factor. 
There can be a variety of SiFs, and lack of consideration of these factors can cause difficulties in classifying spoof and genuine images that share the same SiFs. To avoid this problem, we aim to build a model that learns features that are discriminative for spoof classification, but insensitive to the variations of the SiFs. 

Face anti-spoofing databases have been collected under various acquisition scenarios by varying SiFs, such as identities and camera sensors. In order to consider various SiFs, we consider every possible factor that is provided in the form of labels. Among the different types of the SiFs, at most three types of SiFs are commonly provided as labels from each database: (1) identities of each face, (2) environments (i.e., conditions of illuminations and backgrounds), and (3) sensors (i.e., types of cameras) (Table \ref{db_config}).
In our method, the adversarial learning between an encoder and multiple discrimination heads is conducted to suppress the SiFs.

\subsection{Spoof-irrelevant Factors Suppression}
\label{sif_suppression}
We introduce the first adversarial learning scheme that suppresses spoof-irrelevant factors, so that the trained model becomes invariant to them.
In this learning scheme, an encoder and multiple discrimination heads are adversarially trained, and SiFs are suppressed in the encoded features (Figure \ref{fig:method_overview}).
\smallbreak

\noindent{\textbf{Spoof Classification.}} 
We are given a set $X$ of images, a corresponding set $Y$ of spoof class labels, and a corresponding set $F^{k}$ of labels of each SiFs, where $k$ is a SiF in a set $\mathcal{K} = \{identity, environment, sensor\}$.
Here, the entire spoofing network consists of an encoder $E$ and a spoof classifier $C$ (Figure \ref{fig:architecture}).
Given an image and its corresponding label $(x, y) \sim (X, Y)$, the encoder $E$ and the spoof classifier $C$ are trained to predict whether the image is spoof or genuine; each does this by minimizing the following spoof classification loss:
\begin{multline}
  \mathcal{L}_{cls}(C,E) = -\mathbb{E}_{x,y \sim X,Y}\Big[(1-y)\log (1-\sigma(C(E(x))))\\ 
  + y\log (\sigma(C(E(x))))\Big],
  \label{cls_loss}
\end{multline}
where $\sigma$ denotes a softmax function.
Training the network solely with the spoof classification loss leads to degraded results, and we hypothesize that this degradation is a result of features that are entangled with SiFs.
\smallbreak

\noindent{\textbf{Spoof-irrelevant Factors Suppression.}} 
We introduce a new learning scheme that makes the model focus on the spoof factor by discarding SiFs, so that accurate face anti-spoofing is conducted.
In the learning scheme, we deploy multiple discrimination heads that consist of a SiF-aware intermediate layer $I^{k}$ and a discriminator $D^{k}$ where $k$ corresponds to each of the SiFs in $\mathcal{K}$ (Figure \ref{fig:architecture}).
Each discriminator classifies the corresponding SiF from the encoded features by minimizing SiF classification loss:
\begin{multline}
\mathcal{L}_{sif} (E,I^{k},D^{k}) \\
= -\mathbbm{E}_{x,f^{k} \sim X,F^{k}} \Bigg[ \sum_{n=1}^{N^{k}} \mathbbm{1}_{[n=f^{k}]} \log \sigma{(D^{k}(I^{k}(E(x)))}) \Bigg],
\label{sif_loss}
\end{multline}
where $f^{k}$ is a Sif label that corresponds to $x$, and $\mathbbm{1}_{[n=f^{k}]}$ is an indicator function which equals to $1$ if $n=f^{k}$ and $0$ otherwise. 
$N^{k}$ denotes the number of classes in each type of spoof-irrelevant factor (Table \ref{db_config}).
As an opponent of the discrimination heads, the encoder is adversarially trained to maximize the loss with the objective of suppressing SiFs. This learning scheme is interpreted as the following adversarial procedure:
\begin{equation}
\max_{E}\min_{\mathcal{I}, \mathcal{D}} \sum_{k\in \mathcal{K}} \mathcal{L}_{sif} (E,I^{k},D^{k}),
\label{minmax_sif}
\end{equation}
where $\mathcal{I}$ and $\mathcal{D}$ denotes a set of $I^{k}$ and a set of $D^{k}$ for all SiFs in $\mathcal{K}$, respectively.
The maximization process of the encoder is implemented by using a gradient reversal layer (GRL) \cite{ganin2015unsupervised} that is inserted in between the encoder and each of the discrimination heads (Fig \ref{fig:architecture}).
The GRL acts as an identity function during the forward propagation, and reverses the sign of gradients by multiplying $-1$ during the backpropagation.
If SiFs are abundant in the encoded features, the discriminators can easily distinguish the SiFs by using the encoded features with small error. Hence, for the encoder to maximize the loss, the encoder strives to suppress the SiFs during the feature-encoding process. We observe that as the network is successfully trained, the SiF classification losses gradually diverge; this trend implies that the discriminators fail to classify the SiFs from the encoded features, and that the encoder succeeds in suppressing the SiFs so that the encoded features become invariant to the SiFs.

\subsection{Doubly Adversarial Learning}
The first adversarial learning alleviates the effects of SiFs in the encoded features, however, it may cause the encoder to lose discriminative power since the SiF suppression procedure does not guarantee the features to be discriminative in terms of spoof detection. Also, the overall learning of the encoder can be dominated by the SiF suppression procedure since the encoder has to compete against multiple discrimination heads.
In order to address the problem, we deploy an additional learning strategy.
Along with the spoof classification loss (Eq. \ref{cls_loss}), this learning scheme guides the encoder to learn more discriminative features, and the encoder can be strengthened with intensified spoof factors.
In this learning scheme, the group of the encoder and the secondary classifier is adversarially trained against each of the intermediate layers of the discrimination heads (Figure \ref{fig:method_overview})
\smallbreak

\noindent\textbf{Secondary Spoof Classification.}
We deploy a secondary spoof classifier $S$ after each of the SiF-aware intermediate layers (Figure \ref{fig:architecture}). The objective of the secondary spoof classifier is to minimize the secondary spoof classification loss which is defined as:
\begin{multline}
  \mathcal{L}_{scls}(E,I^{k},S) \\
  = -\mathbb{E}_{x,y \sim X,Y}\Big[ (1-y)\log(1-\sigma(S(I^{k}(E(x))))) \\ 
  + y\log(\sigma(S(I^{k}(E(x))))) \Big].
  \label{scls_loss}
\end{multline}
In contrast, each of the intermediate layers is adversarially trained to maximize this loss with the objective of suppressing the spoof factor.
Thus, this learning scheme is interpreted as the following min-max procedure:
\begin{equation}
\min_{E,S}\max_{\mathcal{I}} \sum_{k\in \mathcal{K}} \mathcal{L}_{scls} (E,I^{k},S).
\end{equation}
By adopting this learning scheme, the encoder is adversarially trained to overcome the suppression of the multiple SiF-aware intermediate layers, so the spoof factor of the encoded features can be intensified, and the encoder is collaboratively \cite{song2018collaborative} trained on varied gradients from each SiF-aware intermediate layer.
\smallbreak

\noindent{\textbf{Doubly Adversarial Learning. }}
The overall learning process involves two different adversarial learning processes, so we call it a Doubly Adversarial Learning.
The full objective is summarized by aggregating the defined losses with corresponding weight terms, $\lambda_{sif}^{k}$, for each $k$ in $\mathcal{K}$, and is written as:
\begin{equation}
\begin{aligned}
\min_{\mathcal{I},\mathcal{D}}\max_{E}\sum_{k\in \mathcal{K}}\lambda_{sif}^{k}\mathcal{L}_{sif} (E,I^{k},D^{k}),
\end{aligned}
\label{step1}
\end{equation}
\begin{equation}
\begin{aligned}
\min_{E,C,S}\max_{\mathcal{I},\mathcal{D}}\mathcal{L}_{cls}(E,C) + \sum_{k\in \mathcal{K}}\mathcal{L}_{scls}(E,S,I^{k},D^{k}),
\end{aligned}
\label{step2}
\end{equation}
where Eq. \ref{step1} is a min-max procedure to suppress SiFs, and Eq. \ref{step2} is a min-max procedure to intensify the spoof factor.

\subsection{Doubly Adversarial Suppression Network}
We propose DASN to realize the introduced learning scheme; it learns complex objectives that are in opposition to each other, so we divide the overall learning scheme into two steps (Figure \ref{fig:architecture}).
\smallbreak

\noindent\textbf{Step 1: Update Encoder and Spoof Classifiers.}
We train the encoder $E$, the spoof classifier $C$, and the secondary spoof classifier $S$.
$E$ and $C$ are updated by minimizing the spoof classification loss (Eq. \ref{cls_loss}).
$E$ is also trained to maximize the SiF classification loss (Eq. \ref{sif_loss}) at the same time.
In this way, $E$ is trained to suppress the SiFs and to encode features that are informative for the spoof classification. To suppress SiFs by maximizing the SiF classification loss, the GRL is inserted in between the encoder and each of the discrimination heads. The discrimination heads are not updated during this step and only pass the gradients to the encoder.
\smallbreak

\noindent\textbf{Step 2: Update Discrimination Heads.}
We train the SiF-aware intermediate layers $I^{k}$ and the discriminators $D^{k}$. 
The SiF-aware intermediate layers and the discriminators are trained to classify SiFs from the encoded features. The SiF-aware intermediate layers are trained to maximize the secondary spoof classification loss at the same time. In this way, the intermediate layers are trained to suppress the spoof factor. To maximize the secondary spoof classification loss, the GRL is inserted in between the secondary spoof classifier and each of the intermediate layers. The secondary spoof classifier is not updated in this step, and only passes the gradients to the SiF-aware intermediate layers.
\smallbreak

\noindent\textbf{Inference Stage.}
The discrimination heads and the secondary spoof classifier are only utilized for the training, hence the proposed DASN does not require any additional computational resources in the inference stage.

\section{Experiments}
\subsection{Datasets and Metrics}
\begin{table}[t]
	\centering	
	\small
	\caption{Statistics of SiFs for training sets of public benchmark datasets; each dataset provides labels of three SiFs that corresponds to images: identity, environment and sensor.}
	\begin{tabular}{c|c|c|c}
		\hline
		Task & \# Identity & \# Environment & \# Sensors \\ \hline\hline
		MSU-MSFD            & 15 & 1  & 2 \\
		CASIA-FASD          & 20 & 1  & 3 \\
		Idiap Replay-Attack & 15 & 2 &  1 \\
		Oulu-NPU            & 20 & 3  & 6 \\
		\hline
	\end{tabular}
	\label{db_config}
\end{table}

We train and evaluate the proposed method using four public benchmark datasets (Table \ref{db_config}): 
\smallbreak

\noindent{\textbf{MSU-MSFD \cite{msu_fasd} (M).}} This dataset contains 280 videos, 120 of which are used for training, 160 of which are used for testing. 
The training set consists of 30 videos of genuine faces, and 90 videos of spoof faces, and the testing set consists of 40 videos of genuine faces and 120 videos of spoof faces.
Two types of sensors were used to record 35 individuals, 15 of which appear in the training set and 20 of which appear in the testing set. Variations of environments are not considered in this dataset. 
\smallbreak

\noindent{\textbf{CASIA-FASD \cite{casia_fasd} (C).}} This dataset contains 600 videos, 240 of which are used for training, and 360 of which are used for testing. 
The training set consists of 60 videos of genuine faces, and 180 videos of spoof faces, and the testing set consists of 90 videos of genuine faces and 270 videos of spoof faces.
50 individuals are included, 20 of whom are included in the training set, and 30 of whom are included in the testing set. 
All individuals were captured in natural scenes, without artificial environment unification. 
To record the individuals, three different sensors were used, they produce low-, normal- and high-quality videos, respectively.
\smallbreak

\noindent{\textbf{Idiap Replay-Attack \cite{replay_attack} (I).}} This dataset contains 360 videos of training set and 480 videos of testing. 
The training set consists of 60 videos of genuine faces, and 300 videos of spoof faces, and the testing set consists of 80 videos of genuine faces and 480 videos of spoof faces.
50 individuals appear in the dataset, and all videos of this dataset were created by using the same type of sensor. The videos were recorded under two different illumination and background setting.
\smallbreak

\noindent{\textbf{Oulu-NPU \cite{oulu_npu} (O)}} This dataset contains 1800 videos of training set and 1800 videos of testing set. 
The training set consists of 360 videos of genuine faces, and 1440 videos of spoof faces, and the testing set consists of 360 videos of genuine faces and 1440 videos of spoof faces.
40 individuals appear in the dataset using 6 different types of sensors. The videos were recorded under three different illumination and background setting.
\smallbreak

\noindent{\textbf{Metrics.}} We evaluate the proposed method by following the same evaluation protocols \cite{shao2020regularized,wang2020cross} of the existing methods; one of the datasets is selected as a testing set, and the remaining three are utilized as training sets. Therefore, four evaluation tasks are possible: {O\&C\&I to M}, {O\&M\&I to C}, {O\&C\&M to I}, and {I\&C\&M to O}. We report the performance of the proposed method by using Area Under the Curve (AUC) and Half Total Error Rate (HTER) = (False Acceptance Rate$ + $False Rejection Rate)$ / 2$ \cite{anjos2011counter}.  
\smallbreak

\subsection{Experimental Setup}
We use all frames of all the videos to train and test our method. 
For each frame, we locate face regions by using a face detector \cite{dsfd}, then the located regions are cropped and resized to $256\times256$.
As a backbone network, we use ResNet-18 \cite{resnet} that is pre-trained on ImageNet \cite{deng2009imagenet}. 
We use Xavier initialization \cite{xavier} to initialize the parameters of layers that are not a part of the pre-trained network. 
We apply global average pooling to the output feature map of the backbone network to obtain vectorized features of 512-dimensionality.
Except for the last fully connected (FC) layers, every other FCs are followed by a ReLU activation function, and the number of their hidden nodes is 512.
We train the network using a constant learning rate of $10^{-5}$ with the Adam optimizer \cite{adam} on a single NVIDIA V100 GPU.
The weight terms $\lambda_{sif}^{k}$ are simply selected by observing the initial loss values to balance the loss values in the similar scales; e.g., we use $0.05$ for $\lambda_{sif}^{identity}$, $0.08$ for $\lambda_{sif}^{environment}$ and $0.08$ for $\lambda_{sif}^{sensor}$ on O\&C\&I to M. We set the size of a mini-batch to 32 on the I\&C\&M to O task, and 64 on the remaining three tasks.

\subsection{Evaluations}

\noindent{\textbf{Comparison Counterparts.}} We compare the performance of our approach with following works: MS\_LBP \cite{maatta2011face}, Binary CNN \cite{yang2014learn}, IDA \cite{msu_fasd}, CT \cite{2016TIFScolortxt}, LBPTOP \cite{2014EJIVPlbptop}, Auxiliary \cite{liu2018learning}, MMD-AAE \cite{2018CVPRdgadv}, MADDG \cite{Shao_2019_CVPR}, RFMetaFAS \cite{shao2020regularized}, CCDD \cite{saha2020domain}, and NAS-FAS \cite{yu2020fas}.
\smallbreak

\noindent{\textbf{Comparison with State-of-the-art Methods.}}
The proposed method shows significant improvements over the baseline, and also shows superior results over the state-of-the-art methods in most of the protocols (Table \ref{tab:3domains}).
For O\&C\&I to M, DASN improves 10.00 percentage points on HTER compared to the baseline; the result is 5.56 percentage points lower than HTER of the state-of-the-art method. 
For I\&C\&M to O, DASN improves 4.13 percentage points on HTER compared to the baseline; the result is 1.39 percentage points lower than HTER of the state-of-the-art method. 
\smallbreak

\noindent{\textbf{Limited Source Domains.}}
We evaluate the proposed method on another task that uses a training set consisting of only two domains (Table \ref{tab:2domains}).
Our DASN also outperforms the other methods on those tasks. 
For M\&I to C, DASN improves 6.77 percentage points on HTER compared to the best model; 
for M\&I to O, DASN improves 12.25 percentage points on HTER compared to the best model.
\\

\begin{table*}[h!]
	\centering	
	\small
	\caption{Evaluation results on four tasks. We compare our baseline and DASN with the state-of-the-art methods of domain generalization for face anti-spoofing. }
	\begin{tabular}{c|c|c|c|c|c|c|c|c}
		\hline
		\multirow{2}{*}{\textbf{Method}} & \multicolumn{2}{c|}{\textbf{O\&C\&I to M}} & \multicolumn{2}{c|}{\textbf{O\&M\&I to C}} 
		& \multicolumn{2}{c|}{\textbf{O\&C\&M to I}}& \multicolumn{2}{c}{\textbf{I\&C\&M to O}}\\ \cline{2-9} 		
		& HTER(\%)& AUC(\%)& HTER(\%)& AUC(\%)& HTER(\%)& AUC(\%)& HTER(\%)& AUC(\%)\\ \hline\hline
		MS\_LBP                             & 29.76 & 78.50	& 54.28 & 44.98	& 50.30 & 51.64 & 50.29 & 49.31\\	
		Binary CNN                          & 29.25 & 82.87 & 34.88 & 71.94 & 34.47 & 65.88 & 29.61 & 77.54\\
		IDA                                 & 66.67 & 27.86	& 55.17 & 39.05	& 28.35 & 78.25	& 54.20 & 44.59\\
		CT                                  & 28.09 & 78.47 & 30.58 & 76.89 & 40.40 & 62.78 & 63.59 & 32.71\\
		LBPTOP                              & 36.90 & 70.80 & 42.60 & 61.05 & 49.45 & 49.54 & 53.15 & 44.09\\		
		Auxiliary (Depth Only)              & 22.72 & 85.88 & 33.52 & 73.15 & 29.14 & 71.69 & 30.17 & 77.61\\
		Auxiliary (All)                     & -     & -     & 28.40 & -     & 27.60 & -     & -     & - \\
		MMD-AAE                             & 27.08 & 83.19 & 44.59 & 58.29	& 31.58 & 75.18	& 40.98 & 63.08\\			
		MADDG                               & 17.69 & 88.06 & 24.50 & 84.51	& 22.19 & 84.99 & 27.98 & 80.02 \\
		PAD-GAN                             & 17.02 & 90.10 & 19.68 & 87.43	& 20.87 & 86.72 & 25.02 & 81.02 \\
		RFMetaFAS                           & 13.89 & 93.98 & 20.27 & 88.16	& 17.30 & 90.48 & 16.45 & 91.16 \\
	    CCDD                                & 15.42 & 91.13 & 17.41 & 90.12 & 15.87 & 91.47 & 14.72  & 93.08 \\
	    NAS-FAS                             & 16.85 & 90.42 & 15.21 & 92.64	& \textbf{11.63} & \textbf{96.98} & 13.16 & 94.18 \cr\hline
		\textbf{Baseline}                   & 18.33 & 92.56 & 22.04 & 89.40 & 19.50 & 80.81 & 15.90 & 90.67 \\
		\textbf{DASN} & \textbf{8.33} & \textbf{96.31} & \textbf{12.04} & \textbf{95.33}	& 13.38 & 86.63&\textbf{11.77}&\textbf{94.65}\\\hline
	\end{tabular}
	\label{tab:3domains}
\end{table*}

\begin{table}[h!]
	\centering	
	\small
	\caption{Evaluation results on another tasks; two different datasets are utilized for training and another dataset is utilized for evaluation.}
	\begin{tabular}{c|c|c|c|c}
		\hline
		\multirow{2}{*}{\textbf{Method}} & \multicolumn{2}{c|}{\textbf{M\&I to C}} & \multicolumn{2}{c}{\textbf{M\&I to O}} \\ \cline{2-5} 		
		&HTER(\%)& AUC(\%)& HTER(\%)& AUC(\%) \\ \hline\hline
			MS\_LBP         & 51.16 &   52.09 & 43.63   &   58.07 \\
			CT              & 55.17 &   46.89 & 53.31   &   45.16 \\
			LBPTOP          & 45.27 &   54.88 & 47.26   &  50.21 \\
			MADDG           & 41.02 &   64.33 & 39.35   &   65.10 \\
		    PAD-GAN         & 31.67 &   75.23 & 34.02   &   72.65 \cr \hline
		\textbf{DASN} & \textbf{21.48} & \textbf{83.41}   & \textbf{21.74} & \textbf{80.87} \\\hline
	\end{tabular}
	\label{tab:2domains}
\end{table}

\begin{table*}[t]
	\centering	
	\small
	\caption{
	    Ablation studies for (1) varying sets of SiFs and (2) varying the proposed adversarial network on four evaluation tasks.
	    Adversarial Suppression Network (ASN) is trained by adopting only the $1^{st}$ adversarial learning; ASN$_d$ is a variation of ASN that uses domain information as a SiF. 
	    $id$, $env$ and $sens$ is an abbreviation of identity, environment and sensor, respectively.} 
	\begin{tabular}{c|c|c|c|c|c|c|c|c|c}
	\hline
	\multirow{2}{*}{\textbf{Method}} & \multirow{2}{*}{\textbf{SiFs}} & \multicolumn{2}{c|}{\textbf{O\&C\&I to M}} & \multicolumn{2}{c|}{\textbf{O\&M\&I to C}} 
	& \multicolumn{2}{c|}{\textbf{O\&C\&M to I}}& \multicolumn{2}{c}{\textbf{I\&C\&M to O}}\\ \cline{3-10} 	 
		& & HTER(\%)&AUC(\%)&HTER(\%)&AUC(\%)&HTER(\%)& AUC(\%)& HTER(\%)&AUC(\%)\\ \hline\hline
		\textbf{Baseline} & $-$         & 18.33 & 92.56 & 22.04 & 89.40 & 19.50 & 80.81 &15.90 & 90.67 \\
		\hline
		\multirow{7}{*}{\textbf{ASN}}   
		& $\{id\}$                      & 13.75 & 93.69 & 15.56 & 91.28 & 14.25 & 82.68 & 13.40 & 93.47 \\
		& $\{env\}$                     & 9.58  & 95.81 & 18.15 & 90.55 & 15.63 & 80.64 & 13.58 & 89.32 \\
		& $\{sens\}$                    & 10.42 & 94.08 & 16.11 & 91.18 & 14.88 & 85.08 & 14.41 & 90.81 \\
		& $\{id,env\}$                  & 9.58  & 95.81 & 17.41 & 92.29 & 17.00 & 86.76 & 14.31 & 91.07 \\
		& $\{id,sens\}$                 & 10.00 & 95.46 & 14.81 & 92.40 & 14.13 & \textbf{87.19}  & 12.92 & 91.63 \\
		& $\{sens,env\}$                & 12.50 & 91.92 & 17.04 & 88.15 & 15.75 & 85.73 & 13.44 & 92.99 \\
		& $\{id,sens,env\}$             & 10.00 & 93.20 & 12.78 & 91.74 & 13.75 & 86.72 & 13.72 & 90.58 \\\hline
		 \textbf{ASN$_d$}& $\{domain\}$ & 8.75 & 96.21  & 13.33 & 92.24 & 14.88 & 82.24 & 14.97 & 91.84 \\\hline
		\textbf{DASN} & $\{id,sens,env\}$ &\textbf{8.33}&\textbf{96.31}&\textbf{12.04}&\textbf{95.33}&\textbf{13.38}&86.63&\textbf{11.77}&\textbf{94.65}\\
		\hline
	\end{tabular}
	\label{tab:ablation}
\end{table*}

\subsection{Ablation Studies}
We demonstrate the effectiveness of the proposed method by conducting extensive ablation studies.
In this ablation studies, all the models also are based on ResNet-18; the baseline model is trained solely with the spoof classification loss. 
\smallbreak

\noindent{\textbf{Spoof-irrelevant Factors Suppression.}}
The performance of models can be significantly improved by suppressing SiFs because these factors can disturb face anti-spoofing (Table \ref{tab:ablation}).
Especially, among the SiFs, it is shown that suppressing the identity factor can be most effective, as the performance improvements over the baseline are largest, except for the O\&C\&I to M task.
\smallbreak

\noindent{\textbf{Doubly Adversarial Learning.}}
We also compare the performance of the models by varying the combinations of the suppressed SiFs (Table \ref{tab:ablation}). We observe that the increase in the number of SiFs does not guarantee the improvements in performance, and some even show degraded results compared to the ones that consider the less number of SiFs.
This observation implies that as the number of the discrimination heads that the encoder competes against increases, the encoder can be disturbed by the discrimination heads, and the spoof factor in the encoded features can be suppressed.
By deploying the doubly adversarial learning scheme, the model stably incorporates the multiple SiFs as the proposed DASN explicitly intensifies the spoof factor in the encoded features, and the collaborative learning \cite{song2018collaborative} behavior also contributes to the improved results.
\smallbreak

\noindent{\textbf{ASN$_d$ vs. DASN.}}
To distinguish our method from the conventional approach of domain generalization \cite{ganin2015unsupervised}, we compare our DASN and ASN$_d$, which uses domain information as a SiF and is trained by the $1^{st}$ adversarial learning scheme (Table \ref{tab:ablation}). DASN shows enhanced generalization ability than ASN$_d$ and this implies that DASN more delicately removes the factors that interfere with the face anti-spoofing.
\smallbreak

\subsection{Discussion}

\begin{figure}[t]
	\begin{center}
		\subfloat[Genuine]{
		    \label{fig:gradcam_real}
			\includegraphics[width=\linewidth]{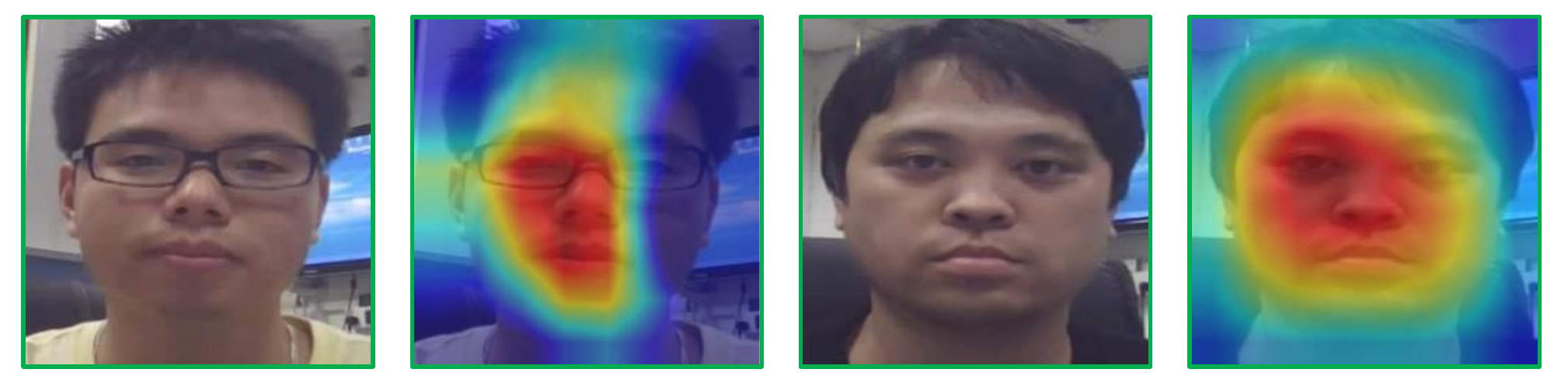}
		}	\\
		\subfloat[Print-Attack]{
		    \label{fig:gradcam_print}
			\includegraphics[width=\linewidth]{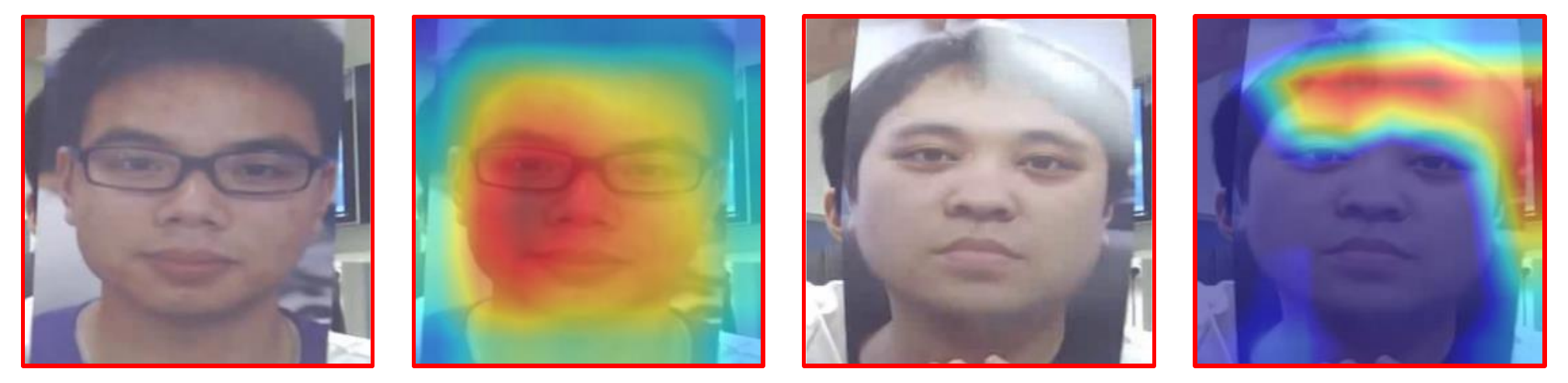}
		}	\\
		\subfloat[Video-Attack]{
		    \label{fig:gradcam_video}
			\includegraphics[width=\linewidth]{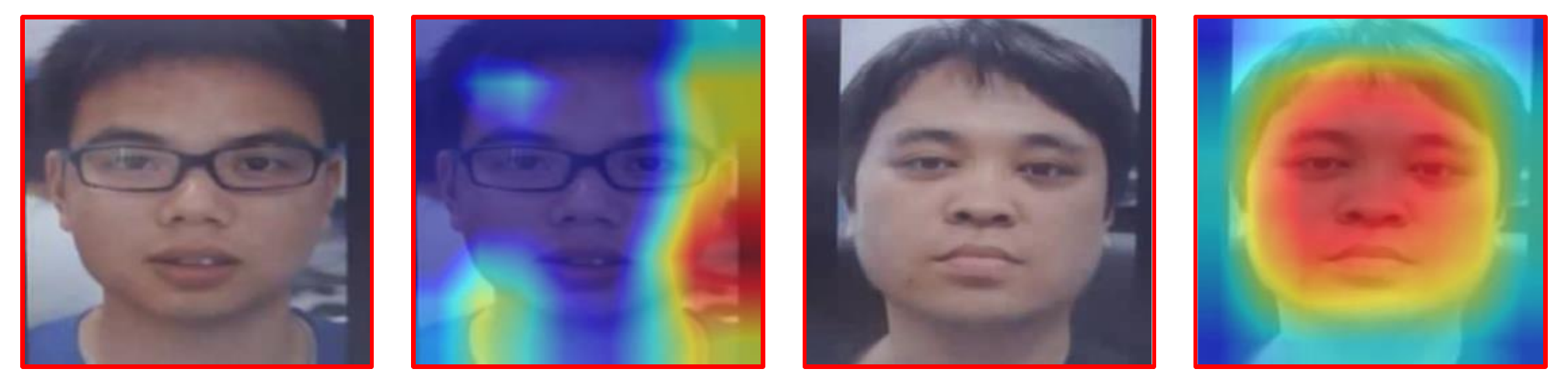}
		}
	\end{center}
	\caption{ 
	Visualization of our method on the evaluation task by Grad-CAM \cite{gradcam} 
	for three classes of inputs:
	(a) genuine, (b) print-attacks and (c) video-attacks.
	The genuine and spoof images share the same SiFs.
  }
	\label{fig:gradcam_result}
\end{figure}

\begin{figure*}[t]
	\begin{center}
		\includegraphics[width=\linewidth]{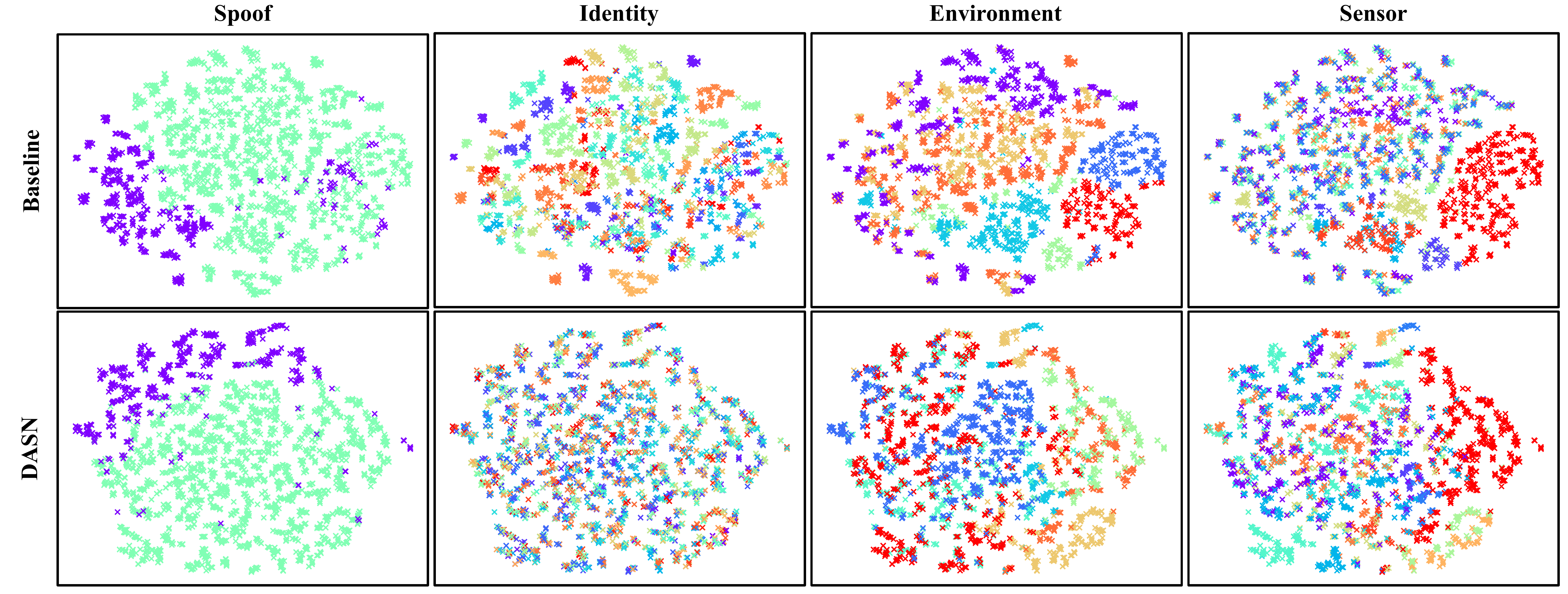}
	\end{center}
	\caption{
	Visualization of our method on the evaluation task by t-SNE \cite{tsne} 
	depending on spoof factor and three SiFs. Each color represents each class of the spoof factor and SiFs.}
	\label{fig:tsne_result}
\end{figure*}

\noindent{\textbf{Grad-CAM Visualization.}} We present a visualization of Gradient-weighted Class Activation Mapping (Grad-CAM) \cite{gradcam} that produces a localization map highlighting important regions for predicting the classes, so that more insights on how the proposed network makes decisions can be obtained (Figure \ref{fig:gradcam_result}). The visualized results show that the attended regions can be diversified, even if the spoof and genuine images share the same SiFs, and this observation supports that our method is insensitive to the SiFs.

In addition, we observe that for spoof images, the attended regions are diversely located in the image, whereas the regions are consistently attended on facial areas for the genuine images. This trend implies that the clues for discriminating spoof images are more diversely located than the clues for the genuine images.
\smallbreak

\noindent{\textbf{t-SNE Visualization.}}
We visualize the distributions of the output features of DASN by using t-SNE \cite{tsne} (Figure \ref{fig:tsne_result}).
It is shown that the output features of DASN are more clearly clustered for the spoof factor.
Compared to the output features baseline model, the output features of the DASN are more dispersed, even though they share the same SiFs, and this trend is most distinguished for identity.
This implies that the proposed method is effective in suppressing the SiFs, and intensifying the spoof factor.
\smallbreak

\section{Conclusion}
In this paper, we propose a DASN that adopts a doubly adversarial learning scheme to improve generalization capability for face anti-spoofing. In our learning scheme, the encoder is trained against multiple discrimination heads to maximize the SiF classification loss, with the objective of suppressing the SiFs in the encoded features. In addition, the encoder is trained to learn the intensified spoof factor by involving additional adversarial learning, where the encoder aims to minimize the secondary spoof classification loss with the objective of overcoming the suppressions of SiF-aware intermediate layers.
The extensive empirical evaluation results on public benchmark datasets demonstrate the effectiveness of our proposed method.

\clearpage

{\small
\bibliographystyle{egbib}
\bibliography{egbib}

\begin{thebibliography}{10}\itemsep=-1pt

\bibitem{anjos2011counter}
Andr{\'e} Anjos and S{\'e}bastien Marcel.
\newblock Counter-measures to photo attacks in face recognition: a public
  database and a baseline.
\newblock In {\em IEEE International Joint Conference on Biometrics}, 2011.

\bibitem{atoum2017face}
Yousef Atoum, Yaojie Liu, Amin Jourabloo, and Xiaoming Liu.
\newblock Face anti-spoofing using patch and depth-based cnns.
\newblock In {\em IEEE International Joint Conference on Biometrics}, 2017.

\bibitem{boulkenafet2015face}
Zinelabidine Boulkenafet, Jukka Komulainen, and Abdenour Hadid.
\newblock Face anti-spoofing based on color texture analysis.
\newblock In {\em IEEE International Conference on Image Processing}, 2015.

\bibitem{2016TIFScolortxt}
Zinelabidine Boulkenafet, Jukka Komulainen, and Abdenour Hadid.
\newblock Face spoofing detection using colour texture analysis.
\newblock In {\em IEEE Transactions on Information Forensics and Security},
  2016.

\bibitem{oulu_npu}
Zinelabinde Boulkenafet, Jukka Komulainen, Lei Li, Xiaoyi Feng, and Abdenour
  Hadid.
\newblock Oulu-npu: A mobile face presentation attack database with real-world
  variations.
\newblock In {\em IEEE International Conference on Automatic Face and Gesture
  Recognition}, 2017.

\bibitem{replay_attack}
Ivana Chingovska, Andr{\'e} Anjos, and S{\'e}bastien Marcel.
\newblock On the effectiveness of local binary patterns in face anti-spoofing.
\newblock In {\em International Conference of the Biometrics Special Interest
  Group}, 2012.

\bibitem{deng2009imagenet}
Jia Deng, Wei Dong, Richard Socher, Li-Jia Li, Kai Li, and Li Fei-Fei.
\newblock Imagenet: A large-scale hierarchical image database.
\newblock In {\em IEEE Conference on Computer Vision and Pattern Recognition},
  2009.

\bibitem{deng2019arcface}
Jiankang Deng, Jia Guo, Niannan Xue, and Stefanos Zafeiriou.
\newblock Arcface: Additive angular margin loss for deep face recognition.
\newblock In {\em IEEE Conference on Computer Vision and Pattern Recognition},
  2019.

\bibitem{ganin2015unsupervised}
Yaroslav Ganin and Victor Lempitsky.
\newblock Unsupervised domain adaptation by backpropagation.
\newblock In {\em International Conference on machine learning}, 2015.

\bibitem{xavier}
Xavier Glorot and Yoshua Bengio.
\newblock Understanding the difficulty of training deep feedforward neural
  networks.
\newblock In {\em International Conference on Artificial Intelligence and
  Statistics}, 2010.

\bibitem{resnet}
Kaiming He, Xiangyu Zhang, Shaoqing Ren, and Jian Sun.
\newblock Deep residual learning for image recognition.
\newblock In {\em IEEE Conference on Computer Vision and Pattern Recognition},
  2016.

\bibitem{kim2019basn}
Taewook Kim, YongHyun Kim, Inhan Kim, and Daijin Kim.
\newblock Basn: Enriching feature representation using bipartite auxiliary
  supervisions for face anti-spoofing.
\newblock In {\em IEEE International Conference on Computer Vision Workshops},
  2019.

\bibitem{Kim_2020_CVPR}
Yonghyun Kim, Wonpyo Park, Myung-Cheol Roh, and Jongju Shin.
\newblock Groupface: Learning latent groups and constructing group-based
  representations for face recognition.
\newblock In {\em IEEE Conference on Computer Vision and Pattern Recognition}.

\bibitem{adam}
Diederik~P Kingma and Jimmy Ba.
\newblock Adam: A method for stochastic optimization.
\newblock In {\em International Conference on Learning Representations}, 2014.

\bibitem{komulainen2013context}
Jukka Komulainen, Abdenour Hadid, and Matti Pietik{\"a}inen.
\newblock Context based face anti-spoofing.
\newblock In {\em IEEE International Conference on Biometrics: Theory,
  Applications and Systems}, 2013.

\bibitem{2018CVPRdgadv}
Haoliang Li, Sinno~Jialin Pan, Shiqi Wang, and Alex~C. Kot.
\newblock Domain generalization with adversarial feature learning.
\newblock In {\em IEEE Conference on Computer Vision and Pattern Recognition},
  2018.

\bibitem{dsfd}
Jian Li, Yabiao Wang, Changan Wang, Ying Tai, Jianjun Qian, Jian Yang, Chengjie
  Wang, Jilin Li, and Feiyue Huang.
\newblock Dsfd: Dual shot face detector.
\newblock In {\em IEEE Conference on Computer Vision and Pattern Recognition},
  2019.

\bibitem{li2016original}
Lei Li, Xiaoyi Feng, Zinelabidine Boulkenafet, Zhaoqiang Xia, Mingming Li, and
  Abdenour Hadid.
\newblock An original face anti-spoofing approach using partial convolutional
  neural network.
\newblock In {\em IEEE International Conference on Image Processing Theory,
  Tools and Applications}, 2016.

\bibitem{liu2018learning}
Yaojie Liu, Amin Jourabloo, and Xiaoming Liu.
\newblock Learning deep models for face anti-spoofing: Binary or auxiliary
  supervision.
\newblock In {\em IEEE Conference on Computer Vision and Pattern Recognition},
  2018.

\bibitem{tsne}
Laurens van~der Maaten and Geoffrey Hinton.
\newblock Visualizing data using t-sne.
\newblock {\em Journal of machine learning research}, 2008.

\bibitem{maatta2011face}
Jukka M{\"a}{\"a}tt{\"a}, Abdenour Hadid, and Matti Pietik{\"a}inen.
\newblock Face spoofing detection from single images using micro-texture
  analysis.
\newblock In {\em IEEE International Joint Conference on Biometrics}, 2011.

\bibitem{patel2016cross}
Keyurkumar Patel, Hu Han, and Anil~K Jain.
\newblock Cross-database face antispoofing with robust feature representation.
\newblock In {\em Chinese Conference on Biometric Recognition}, 2016.

\bibitem{2014EJIVPlbptop}
Tiago~Freitas Pereira and et al.
\newblock Face liveness detection using dynamic texture.
\newblock In {\em EURASIP Journal on Image and Video Processing}, 2014.

\bibitem{saha2020domain}
Suman Saha, Wenhao Xu, Menelaos Kanakis, Stamatios Georgoulis, Yuhua Chen,
  Danda Pani~Paudel, and Luc Van~Gool.
\newblock Domain agnostic feature learning for image and video based face
  anti-spoofing.
\newblock In {\em IEEE Conference on Computer Vision and Pattern Recognition
  Workshops}, 2020.

\bibitem{gradcam}
Ramprasaath~R Selvaraju, Michael Cogswell, Abhishek Das, Ramakrishna Vedantam,
  Devi Parikh, and Dhruv Batra.
\newblock Grad-cam: Visual explanations from deep networks via gradient-based
  localization.
\newblock In {\em IEEE International Conference on Computer Vision}, 2017.

\bibitem{Shao_2019_CVPR}
Rui Shao, Xiangyuan Lan, Jiawei Li, and Pong~C. Yuen.
\newblock Multi-adversarial discriminative deep domain generalization for face
  presentation attack detection.
\newblock In {\em IEEE Conference on Computer Vision and Pattern Recognition},
  2019.

\bibitem{shao2020regularized}
Rui Shao, Xiangyuan Lan, and Pong~C Yuen.
\newblock Regularized fine-grained meta face anti-spoofing.
\newblock In {\em AAAI Conference on Artificial Intelligence}, 2020.

\bibitem{song2018collaborative}
Guocong Song and Wei Chai.
\newblock Collaborative learning for deep neural networks.
\newblock In {\em International Conference on Neural Information Processing
  Systems}, 2018.

\bibitem{Stehouwer_2020_CVPR}
Joel Stehouwer, Amin Jourabloo, Yaojie Liu, and Xiaoming Liu.
\newblock Noise modeling, synthesis and classification for generic object
  anti-spoofing.
\newblock In {\em IEEE Conference on Computer Vision and Pattern Recognition},
  2020.

\bibitem{wang2020cross}
Guoqing Wang, Hu Han, Shiguang Shan, and Xilin Chen.
\newblock Cross-domain face presentation attack detection via multi-domain
  disentangled representation learning.
\newblock In {\em IEEE Conference on Computer Vision and Pattern Recognition},
  2020.

\bibitem{cosface}
Hao Wang, Yitong Wang, Zheng Zhou, Xing Ji, Dihong Gong, Jingchao Zhou, Zhifeng
  Li, and Wei Liu.
\newblock Cosface: Large margin cosine loss for deep face recognition.
\newblock In {\em IEEE Conference on Computer Vision and Pattern Recognition},
  2018.

\bibitem{msu_fasd}
Di Wen, Hu Han, and Anil~K Jain.
\newblock Face spoof detection with image distortion analysis.
\newblock {\em IEEE Transactions on Information Forensics and Security}, 2015.

\bibitem{yang2014learn}
Jianwei Yang, Zhen Lei, and Stan~Z Li.
\newblock Learn convolutional neural network for face anti-spoofing.
\newblock {\em arXiv preprint arXiv:1408.5601}, 2014.

\bibitem{yang2013face}
Jianwei Yang, Zhen Lei, Shengcai Liao, and Stan~Z Li.
\newblock Face liveness detection with component dependent descriptor.
\newblock In {\em IAPR International Conference on Biometrics}, 2013.

\bibitem{yu2020fas}
Zitong Yu, Jun Wan, Yunxiao Qin, Xiaobai Li, Stan~Z Li, and Guoying Zhao.
\newblock Nas-fas: Static-dynamic central difference network search for face
  anti-spoofing.
\newblock {\em IEEE Transactions on Pattern Analysis and Machine Intelligence},
  2020.

\bibitem{casia_fasd}
Zhiwei Zhang, Junjie Yan, Sifei Liu, Zhen Lei, Dong Yi, and Stan~Z Li.
\newblock A face antispoofing database with diverse attacks.
\newblock In {\em IAPR International Conference on Biometrics}, 2012.

\bibitem{regularface}
Kai Zhao, Jingyi Xu, and Ming-Ming Cheng.
\newblock Regularface: Deep face recognition via exclusive regularization.
\newblock In {\em IEEE Conference on Computer Vision and Pattern Recognition},
  2019.

\end{thebibliography}
}

\end{document}